\documentclass[UTF8,lettersize,journal]{IEEEtran}
\usepackage{amsmath,amsfonts,amssymb}
\usepackage{algorithmic}
\usepackage{algorithm}
\usepackage{array}
\usepackage[caption=false,font=normalsize,labelfont=sf,textfont=sf]{subfig}
\usepackage{textcomp}
\usepackage{stfloats}
\usepackage{url}
\usepackage{verbatim}
\usepackage{graphicx}
\usepackage{cite}
\usepackage{multirow}
\usepackage{booktabs}
\usepackage[justification=centering]{caption}
\usepackage{hyperref}
\usepackage{color}

\begin{document}

\title{Building Change Detection in Earthquake: A Multi-Scale Interaction Network and A Change Detection Dataset}

\author{Yunlong Liu$^{*}$, Zekai Zhang
	% <-this % stops a space
	% <-this % stops a space
	\thanks{$^{*}$ Corresponding author. Y. Liu and Z. Zhang are with the School of Control Science and Engineering, Shandong University, Ji'nan 250061, China (e-mail: xxlnova@163.com, 202420810@mail.sdu.edu.cn).
}}

% The paper headers
\markboth{Journal of \LaTeX\ Class Files,~Vol.~14, No.~8, August~2021}%
{Shell \MakeLowercase{\textit{et al.}}: A Sample Article Using IEEEtran.cls for IEEE Journals}

%\IEEEpubid{0000--0000/00\$00.00~\copyright~2021 IEEE}
% Remember, if you use this you must call \IEEEpubidadjcol in the second
% column for its text to clear the IEEEpubid mark.

\maketitle

\begin{abstract}
As one of the most destructive natural disasters, earthquakes have struck many countries around the world in recent years, causing serious economic losses. Change detection (CD) can be applied to post-earthquake damage assessment as it can infer destroyed change regions from multi-temporal remote sensing images. Furthermore, the CD with short imaging interval will better satisfy the needs of the emergency rescues after earthquakes. However, the capability of current methods built on deep neural networks is limited because the dataset with short imaging interval is absent. To meet post-disaster immediate relief, we create a CD dataset, Turkey earthquake CD dataset (TUE-CD), for the evaluation of building damage in the short term after an earthquake. Because of the short acquisition interval of the post-event images, the imaging angle is different for different temporal images, which leads to some side-looking problems. To deal with these challenges, we present a multi-scale feature interaction network (MSI-Net) for efficient interaction between bi-temporal features, as well as mitigating the effect of side-looking problems. Specifically, the proposed MSI-Net consists of joint cross-attention (JCA) modules, multi-scale offset calibration (MOC) modules, and feature integration (FeI) modules. The JCA module unifies channel cross-attention and spatial joint attention for sufficient feature interaction. The MOC module further estimates the offsets to align the bi-temporal image with the multi-scale features. Finally, calibrated features and multi-scale features are fused by FeI modules for the prediction of changed areas. Experiments on the WHU-CD, CLCD, and the constructed TUE-CD dataset indicate that the proposed MSI-Net provides better results than considered state-of-the-art CD methods.  
\end{abstract}


\section{Introduction}
\IEEEPARstart{I}{n} recent years, earthquake disasters have occurred frequently around the globe, which lead to massive building collapses. The collapses are the main cause of human casualties. Therefore, after disaster strikes, it is very essential for appropriate assistance and rescue missions to assess building damage in earthquakes quickly and reliably. Fortunately, change detection (CD) can identify relevant change areas by comparing remote sensing (RS) images of the same location at different times. Owing to its effectiveness, CD has attracted considerable attention and has been applied to many tasks, including natural disaster assessment \cite{Blpher01} , biomass prediction \cite{Clpher06}, and urban planning \cite{Blpher02}. Naturally, CD is used for evaluating post-earthquake building damage \cite{Blpher04, Blpher05}.

At present, deep neural networks (DNNs)-based methods have become the mainstream in the CD task and achieved impressive performance due to its powerful representation capabilities. For example, researchers generally employ convolutional neural networks (CNNs) to extract representational features from bi-temporal RS images to develop the accuracy of CD \cite{Blpher06, Blpher07, Blpher08}. Noman \emph{et al.}\cite{Blpher09} analyzed the semantic change relationships between images by applying deep convolution, which improved the robustness of the network while reducing the computational complexity. SASiamNet \cite{Blpher10} adaptively assigned and integrated the discriminative information and improves the network's ability to recognize fuzzy and small change targets. Fang \emph{et al.}\cite{Blpher11} directly exchanged bi-temporal features in the spatial and channel dimensions, which facilitated the exchange of information between features and boosted the change detection performance of the network. Wang \emph{et al.}\cite{Blpher12} constructed a weak label generation method based on image blocks and enhanced the capability of the network to recognize foreground information via an adaptive weighting mechanism. Although CNNs perform well in terms of feature extraction, limited receptive fields result in their inability to model global relationships of objects.

In order to make DNNs focus on the changed regions, attention mechanisms are considered. For example, Ji \emph{et al.} \cite{Blpher13} fully exploited the difference information between images through an interactive differential attention mechanism to mitigate the affect of irrelevant information on CD accuracy. Yang \emph{et al.} \cite{Blpher14} developed hyperbolic similarity attention mechanisms to map depth features to non-Euclidean spaces and explored the potential effects of hyperbolic information in on CD. In CSI-Net \cite{Clpher09}, the style prior between bi-temporal RS images is extracted for suppressing spectral differences. MVHFNet \cite{Blpher22} leveraged the powerful global spatial modelling capabilities of hypergraphs to capture complex higher-order relationships from multiple views. Transformer can establish long term dependencies between individual blocks in an image and received attention within the field of CD. Bit-CD\cite{Blpher15} combines the strengths of CNN and transformer to explore global semantic information in extracted features while highlighting the change regions. Noman \emph{et al.} \cite{Blpher16} introduced a new shuffled sparse attention mechanism to replace the traditional self-attention mechanism in transformer, and achieved competitive CD performance without pre-training on other datasets. In CTD-Former\cite{Blpher17} the cross-temporal difference (CTD) attention was designed to capture the intrinsic relationship differences in bi-temporal RS images. Yu \emph{et al.} \cite{Alpher18} proposed context-aware relative position encoding, which improves the global relationship modeling capability of the transformer and facilitates the network in capturing the small change regions in the dispersion.
\begin{figure}[ht!]
	\centering
	\includegraphics[scale=0.3]{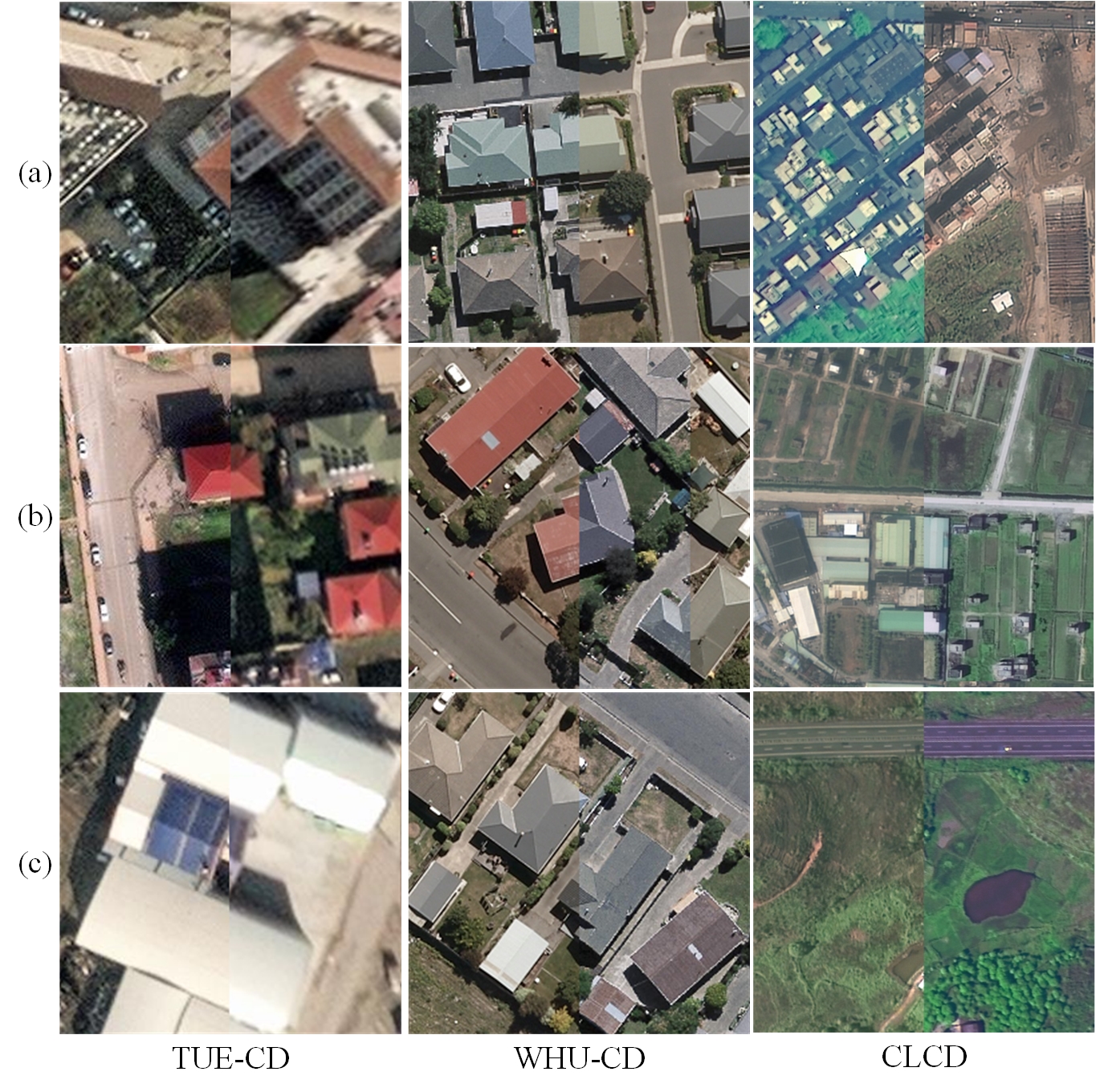}
	\captionsetup{font={footnotesize}}
	\caption{Mismatch problem appear in the TUE-CD, WHU-CD, CLCD datasets.}
	\label{Fig.1}
	\vspace{-0.2cm}
\end{figure}
Although DNN-based CD methods have been explored extensively, most of them cannot produce good results in terms of the estimation task of building damage induced by earthquake when the post-event images are immediately obtained after earthquake. Obviously, the CD with short imaging interval after a earthquake can better satisfy the needs of the emergency rescues after earthquakes. However, there is a long interval between the time of earthquakes and the	imaging time of post-event data for building damage estimation. For example, the acquisition interval of the earthquake data in the xBD dataset \cite{Blpher18} varies from a few months to a few years on average. Therefore, we label a novel CD dataset for the assessment of construction damage in earthquakes, the Turkey Earthquake change detection dataset (TUE-CD), from the images provided by the open data project of Maxar. The bi-temporal RS images in TUE-CD are both collected by WorldView-2 and record the devastating magnitude 7.8 earthquake striking Turkey on Monday, February 6, 2023. In TUE-CD dataset, the post-event data is acquired within 5 days after earthquakes. However, because of the short interval between acquisition time, there are some differences in terms of imaging angels of bi-temporal RS images, which lead to side-looking issues, especially in areas containing high buildings. Specifically, Fig. 1(a) shows the samples in our built TUE-CD dataset. It can be observed that there are some mismatches of buildings caused by side-looking. Besides, similar phenomena also appear in existing datasets as shown in Figs. 1(b) and 1(c), which are ignored by most of CD methods.

To address the issues mentioned above, we suggest a multi-scale feature interaction network with offset calibration (MSI-Net) to efficiently explore the change discrimination information between bi-temporal RS images. Specifically, we employ a ResNet-18 \cite{Clpher07}  as the encoder to extract features of different scale. Then, a joint cross-attention module (JCA) is deigned for feature interaction of these multi-scale features. In the JCA module, multiple attention mechanisms along spatial and channel dimensions are computed for sufficient interaction, which can guide the proposed MSI-Net to focus on changed regions. Then, we build a multi-scale offset calibration module (MOC) to further align the multi-scale features of changed areas. Through this module,  the mismatch effects between bi-temporal RS images caused by side-looking issue are alleviated. Next, the features calibrated by the MOC module is integrated with the multi-scale features through the feature integration (FeI) module for better compensation. Finally, the compensated features are combined and progressively up-sampled to generate the change map. Experimental results on WHU-CD, CLCD, and our constructed TUE-CD dataset indicate that the proposed MSI-Net generates competitive results. The contributions of this work are as follows:

1) We create a CD dataset called TUE-CD that can be used to estimate building damage caused by earthquakes. This dataset includes 1656 pairs of bi-temporal RS images with the size of 256×256 captured by the WorldView-2 satellite. There exist mismatches caused by side-looking because the post-event data in this dataset is collected within 5 days after earthquakes.

2) We propose a MSI-Net to suppress the impact of side-looking issues, which consists of JCA, MOC, and FeI modules for better alignment of features from different temporal RS images.

3) We design JCA and MOC modules for sufficient interaction and accurate calibration of bi-temporal features. In the JCA module, the interaction of features realised by the integration of spatial joint attention and channel cross-attention. The offset estimation in the MOC module is devised to mitigate the spatial mismatches.   

To validate the efficiency of the proposed MSI-Net on CD tasks, we conduct intensive experiments on the WHU-CD, CLCD, and the labelled TUE-CD datasets. Compared with nine state-of-art DL-based CD methods, the proposed MSI-Net attains better performance in both visual and objective evaluation. The content of the rest of this paper is organised as follows. In Section II we provide a brief review of traditional CD methods as well as the methods based on DL. In Section III, the overall structure of the proposed MSI-Net and the designed JCA, MOC, and FeI modules are discussed in detail. In Section IV, we execute in-depth experiments and analyse the results. Finally, in section V, the article is summarized.

\section{Related Work}
\subsection{Traditional CD Methods}
Early RS image CD methods are usually algebra-based methods, such as image difference \cite{Blpher23} and image ratio methods \cite{Blpher24}, which directly perform subtraction or division on bi-temporal images. Change vector analysis (CVA) \cite{Blpher25} and its extended variant compressed change vector analysis (C$^{2}$VA) \cite{Clpher03} are performed by comparing pixel values with change vectors, the intensity of which represents the probability of change. Zhang \emph{et al.}\cite{Blpher26} extracted representative pixels using a low-rank representation method and performed CD by principal component analysis (PCA). Although these methods have been intensively investigated, their limited robustness and high dependence on handcrafted features lead to poor performance in complex scenarios.

\subsection{DNN-Based CD Methods}
Considering the advantages of DNN techniques in terms of learning ability, researchers have proposed numerous DNN-based algorithms to detect the change regions between RS images. For example, Yang \emph{et al.}\cite{Blpher27} utilized residual convolution instead of vanilla convolution to accelerate the training of the network and prevent overfitting and network degradation. Ye \emph{et al.}\cite{Blpher28} innovatively combined bi-temporal images on the time dimension and implemented efficient information fusion while extracting features by leveraging the essential properties of 3D convolution. Liu \emph{et al.}\cite{Blpher29} converted the CD project to a image segments problem, as well as replaced the vanilla convolution with deep separable convolution operation, promoting the network more parameter friendly and improving the CD accuracy. A unsupervised CD framework for RS image time series \cite{Clpher01} was designed by leveraging contrast learning and feature tracking. Moreover, this method also can be applied to multiple sensors. Zhang \emph{et al.}\cite{Blpher38} devised a multi-scale registration  algorithm based on differential flow fields to realize spatial-level feature alignment. Chen \emph{et al.} \cite{Clpher04} mitigated the influence of seasonal changes on CD results by introducing a simple but efficient uncertainty mechanism.\cite{1,2,3,4,5,6,7,8,9,10,11,12,13,zhang2026novel,zhang2023idd,zhang2025zero,zhang2024representation,zhang2026unification}

In order to make DNNs focus on the real changed regions, the attention mechanism caught the interest of researchers. For example, Zhai \emph{et al.} \cite{Blpher30}employed contrast attention to reduce the impact of noise in the image on CD and to facilitate the fusion of different temporal features. Zhao \emph{et al.}\cite{Blpher31} explored the potential of weakly supervised learning in CD by generating pseudo-labels with the class attention map. DARNet \cite{Blpher32} employed hybrid attention to generate discriminative features from context information. CIT\cite{Blpher33} further combined adversarial learning and hybrid attention to perform cross-domain translation, by which the style differences between bi-temporal images are suppressed. In CADRL \cite{Blpher40} multiple cascaded cross-attention modules was introduced to explore the difference information between bi-temporal RS images more efficiently. As a representative method of global self-attention, transformer-based CD methods have achieved competitive results \cite{Blpher34, Blpher35,Blpher36}. For instance, AdaptFormer \cite{Blpher37} can adopt adaptive computation strategies for features with different semantic depths, solving the problem of information degradation. Lin \emph{et al.}\cite{Blpher39} analyzed the similarity relationship between two global pixels with the transformer and improved the robustness of CD. SCanNet \cite{Clpher02} simulated the ``from-to" semantic transformations between bi-temporal RS images by a semantic change converter consisting of transformer, which effectively improved the performance of semantic CD.

Generative models are also an important part of DNNs, especially in the image synthesis field. SDACD \cite{Clpher05} constructed easy-to-insert domain-adaptive networks with generative adversarial networks (GANs) to mitigate the impact of style differences on CD. Diffusion model is an emerging generative technique with the advantage of modeling complex data distributions without adversarial training. For example, Ddpm-cd \cite{Blpher42} took a pre-trained diffusion model as the encoder, and designed a lightweight CD decoder, by which the training time was greatly reduced. DGDM \cite{Blpher43} utilized the pre-training knowledge in the text-to-image diffusion model to explore the possibilities of the diffusion model in the CD task.

\begin{figure*}[ht]
	\centering
	\includegraphics[scale=0.4]{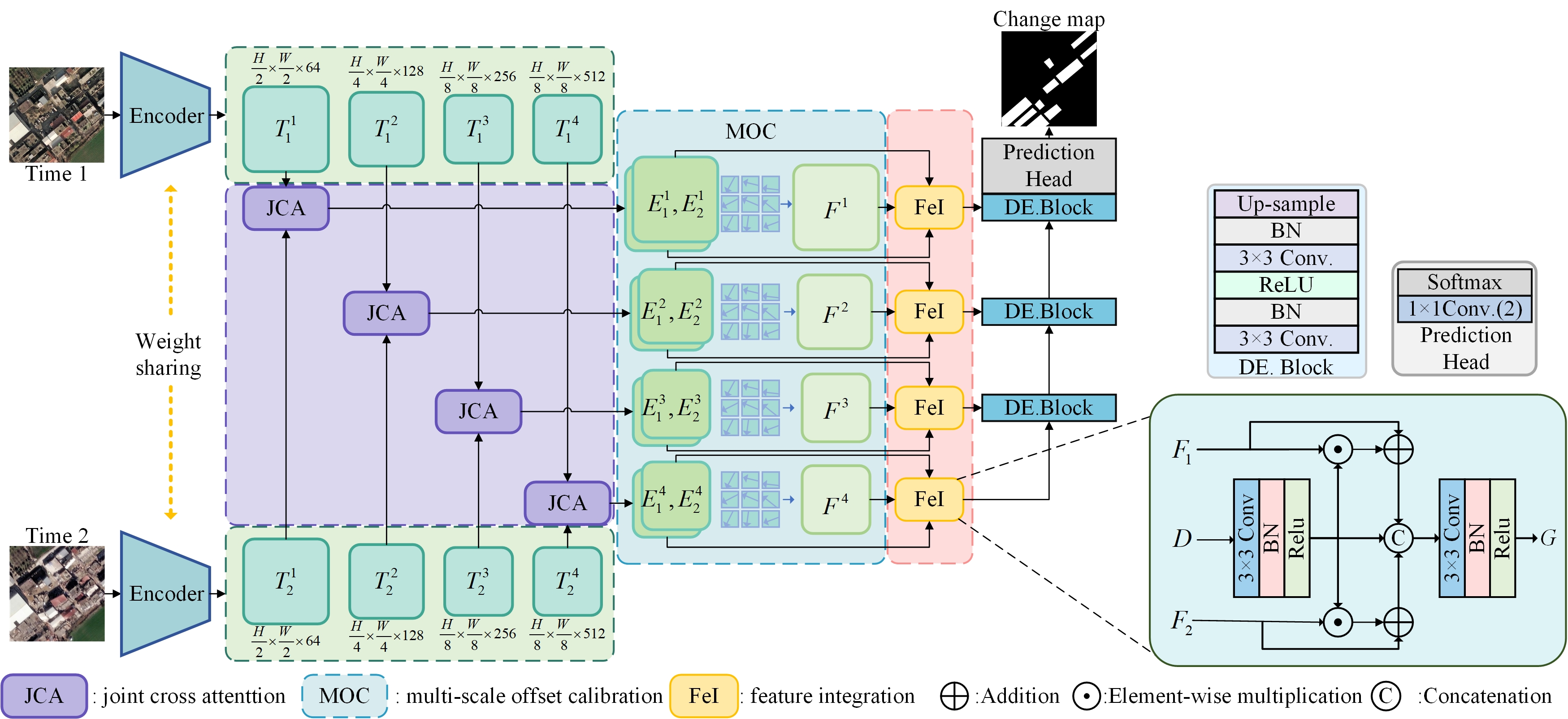}
	\captionsetup{font={footnotesize}}
	\caption{Overall architecture of the proposed MSI-Net}
	\label{Fig.2}
\end{figure*}

\section{MSI-Net}
In this section, we first describe the general architecture of the proposed MSI-Net and the implementation process. Next, we elaborate the structures of the JCA module, the MOC module, and the FeI module. Finally, we present the optimization for the training of the MSI-Net.

\subsection{Overall Architecture}

Fig. 2 demonstrates the overall architecture of the proposed MSI-Net and the specific processes. To begin with, the bi-temporal RS images ${I_1} \in {\mathbb{R}^{H \times W \times 3}}$ and ${I_2} \in {\mathbb{R}^{H \times W \times 3}}$ are processed by a weight-sharing Siamese encoder to extract features with different scales, where the height and width of the input image is $H$ and $W$, respectively, and the number of input channels is three bands, R, G, B. In MSI-Net we eliminate the max pooling operation as well as the linear layer of RenNet-18 and then utilize it as the Siamese feature extractor. Since the shallow features $T_1^1$ and $T_2^1$ extracted by encoders contain less semantic information \cite{Clpher08}, they are directly fed into the MOC module without the interaction of the JCA module. Then, the features $T_1^i, T_2^i \in {\mathbb{R}^{{h_i} \times {w_i} \times {c_i}}},i \in {2,3,4}$ from different images interact with each other by JCA modules. Here, $T_1^i$ and $T_2^i$ represent output features of the $i$-th scale in encoders. Through JCA modules, inter-temporal features are interacted comprehensively, and the semantic gaps between these features are reduced. Subsequently, MOC modules are implemented on the outputs of JCA modules to alleviate the mismatches caused by side-looking issues. Next, the features calibrated by MOC modules are further combined with the outputs of JCA modules for information compensation. Finally, all features from different scales are considered for CD prediction by upsampling, and then the output values are mapped to between 0 and 1 by a prediction head.

\subsection{Joint Cross Attention Module}
Fig. 3 shows the proposed JCA module, which includes two components: the channel cross-attention (CCA) block and the spatial joint attention (SJA) block. In the JCA module, the CCA block is designed to enhance the information exchange of inter-temporal features and make feature distributions more similar. In the SJA block, both cross-attention and self-attention mechanisms are exploited to explore the spatial detail distribution and capture the long-range dependencies among images. CCA and SJA blocks are detailed as follows.

\begin{figure*}[ht]
	\centering
	\includegraphics[scale=0.6]{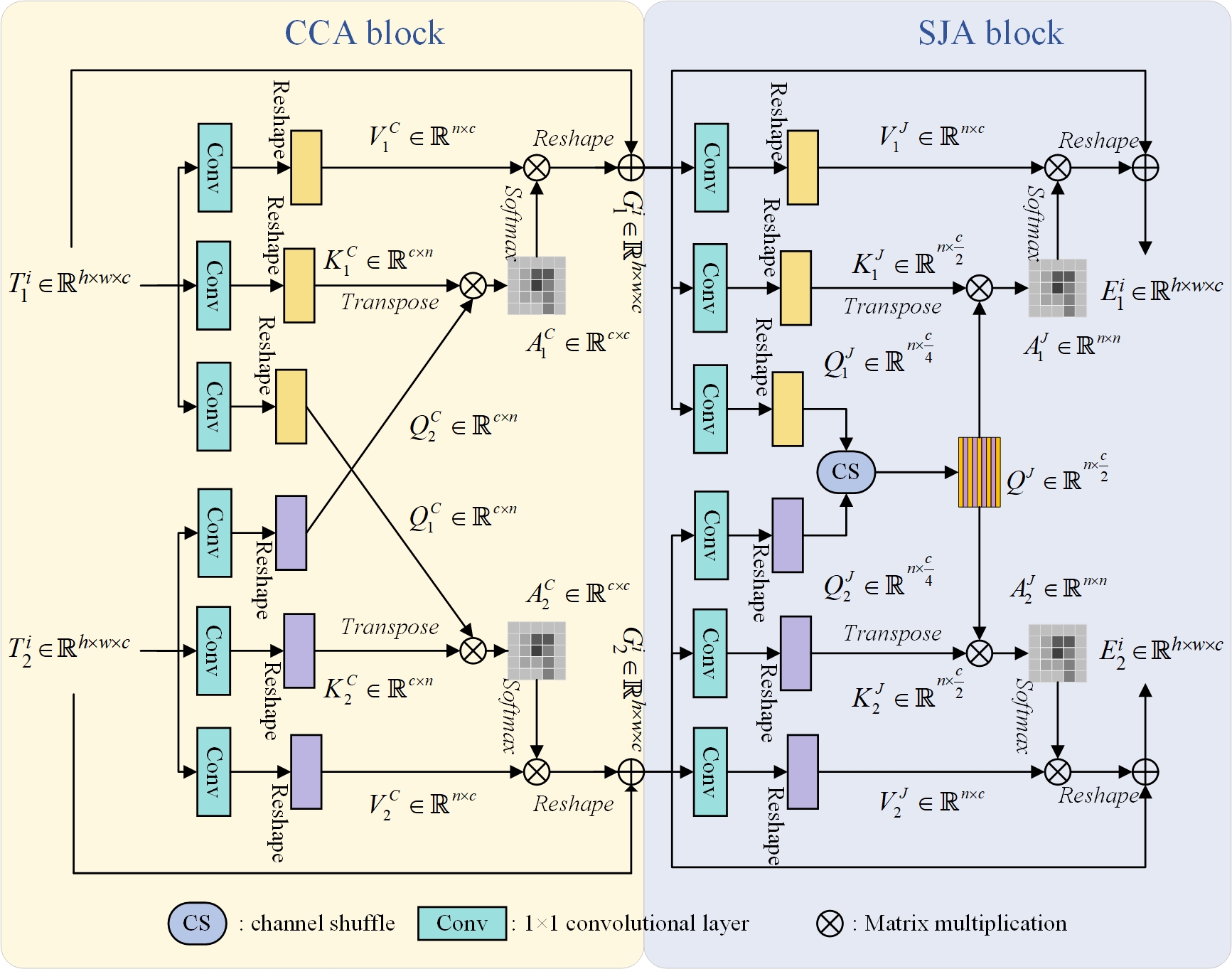}
	\captionsetup{font={footnotesize}}
	\caption{ Architecture of JCA module.}
	\label{fig.3}
\end{figure*}

\textbf{\textit{CCA Block}}: In this block, the features with the dimensions ${h \times w \times c}$ from bi-temporal images are first projected by three 1×1 convolution layers to obtain corresponding queries, keys, and values. Then, the queries $Q_{1}^{C} \in {\mathbb{R}^{c \times n}}$ and $Q_{2}^{C} \in {\mathbb{R}^{c \times n}}$ are exchanged for the calculation of cross-temporal attention. $n$ is equal to $n=h \times w$. Taking $T_{1}^i \in {\mathbb{R}^{h \times w \times c}}$ from the pre-event image as an example, the query $Q_{2}^{C}$ from $T_{2}^i \in {\mathbb{R}^{h \times w \times c}}$ is multiplied with the transpose of key $K_{1}^{C} \in {\mathbb{R}^{c \times n}}$ to obtain the cross-temporal attention map $A_{1}^{C} \in {\mathbb{R}^{c \times c}}$. Then, $A_{1}^{C}$ is normalized by Softmax, in which the similarity among different feature channels are encoded. Finally, the value $V_{1}^{C} \in {\mathbb{R}^{c \times n}}$ is combined with the cross-temporal attention to exchange the content information from different temporal images, and the reshaped cross-temporal feature is further injected into $T_{1}^i$. Specifically, the CCA of $T_{1}^i$ is defined as:
\begin{equation}
	\begin{aligned}
		\begin{array}{c}
			\begin{array}{l}
				G_1^i = T_1^i + {\cal R}\left( {V_1^CSoftmax\left( {Q_2^C{\cal T}\left( {K_1^C} \right)} \right)} \right)
			\end{array}
		\end{array}
	\end{aligned}
\end{equation}
where ${\cal R}$ and ${\cal T}$ denote the reshaping and transpose operators, respectively. Similarly, the CCA of $T_{2}^i$ also can be obtained by integrating the query $Q_{1}^{C}$ from $T_{1}^i$. In this way, the cross-temporal interaction in terms of channel information is achieved.
\begin{figure}[ht!]
	\centering
	\includegraphics[scale=0.85]{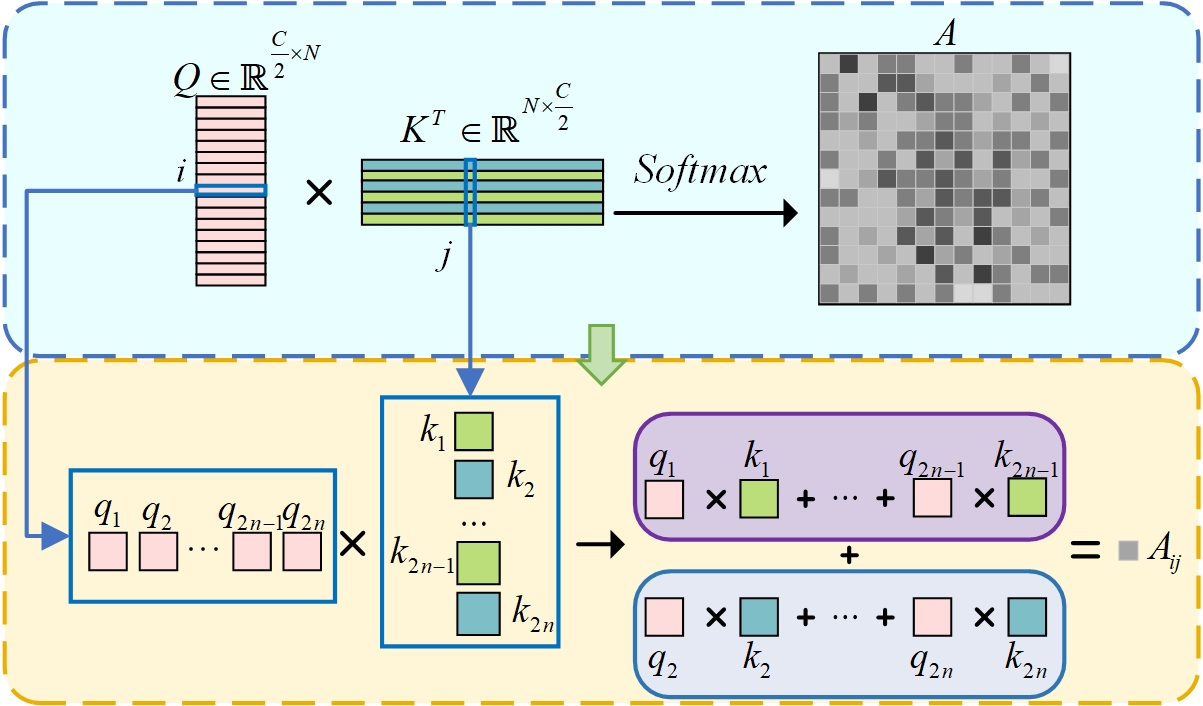}
	\captionsetup{font={footnotesize}}
	\caption{ Procedure of SJA block.}
	\label{fig.4}
\end{figure}
\begin{figure*}[ht!]
	\centering
	\includegraphics[scale=0.7]{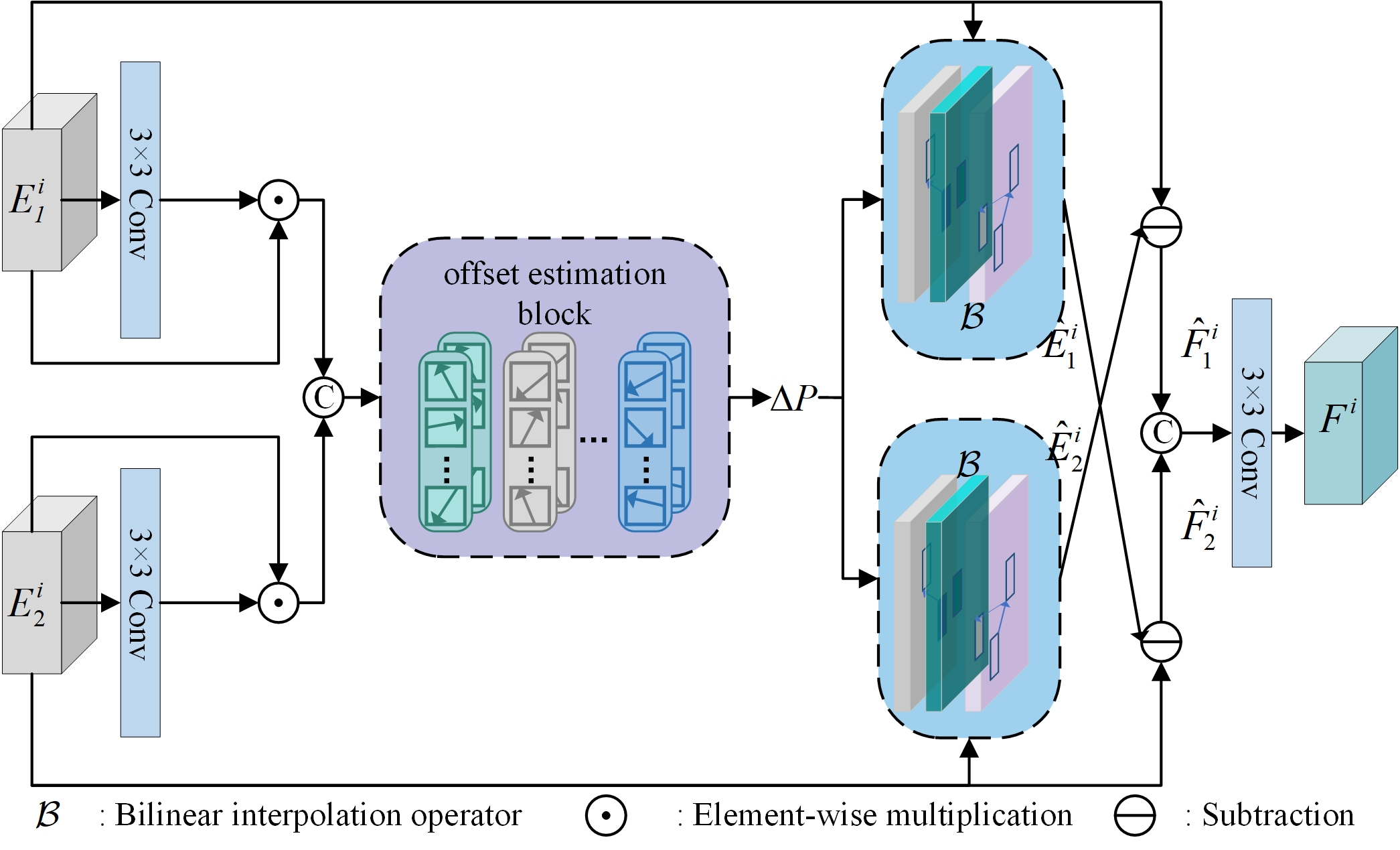}
	\captionsetup{font={footnotesize}}
	\caption{Architecture of MOC module.}
	\label{fig.5}
\end{figure*}

\textbf{\textit{SJA Block}}: In the SJA block, the outputs ${G_{1}} \in {\mathbb{R}^{h \times w \times c}}$ and ${G_{2}} \in {\mathbb{R}^{h \times w \times c}}$ of the CCA block are also processed by three 1×1 convolutional layers and reshaped as embedded queries, keys, and values. Here, the numbers of channels of queries and keys are reduced to $c/4$ and $c/2$ to alleviate the high computational complexity. In this block, the queries $Q_{1}^{J} \in {\mathbb{R}^{\frac{c}{4} \times n}}$ and $Q_{2}^{J} \in {\mathbb{R}^{\frac{c}{4} \times n}}$ of ${G_{1}}$ and ${G_{2}}$ are concatenated as $Q^{J} \in {\mathbb{R}^{\frac{c}{2} \times n}}$ by the channel shuffle operation. Then, the key $K_{1}^{J} \in {\mathbb{R}^{\frac{c}{2} \times n}}$ of ${G_{1}}$ is combined with $Q^{J}$ and then normalized by Softmax to obtain the joint attention map $A_1^{J} \in {\mathbb{R}^{n \times n}}$. In order to embed the joint attention more efficiently, the combination of $K_{1}^{J}$ and $Q^{J}$ according to the formulation in Fig. 4. By the multiplication in Fig. 4, the self-attention of ${G_{1}}$ and cross-attention between ${G_{1}}$ and ${G_{2}}$ are achieved simultaneously. Finally, the joint attention $A_1^{J}$ is implemented on $V_{1}^{joint} \in {\mathbb{R}^{c \times n}}$ for attention embedding. The estimation of SJA of ${G_{1}}$ is summarized as:
\begin{equation}
	\begin{aligned}
		\begin{array}{l}
			\begin{array}{l}
				E_1^i = G_1^i + {\cal R}\left( {Softmax\left( {{Q^J}{\cal T}\left( {K_1^J} \right)} \right)V_1^J} \right)
			\end{array}
		\end{array}
	\end{aligned}
\end{equation}

Through the SJA in this block, the long-range global information in images is captured and the spatial relationships between bi-temporal images are also interacted efficiently.

\subsection{Multi-Scale Offset Calibration Module}
Although the multi-temporal information is sufficiently interacted by the JCA module, the mismatches caused by side-looking are not effectively mitigated. To reduce the influences of side-looking, we suggest a MOC module as shown in Fig. 5. In the MOC module, the features $E_1^i$ and $E_2^i$ from bi-temporal images are first convolved and concatenated for the estimation of offset field. Then, $E_1^i$ and $E_2^i$ are calibrated according to the offset field for better matches of spatial information. Finally, the difference information of calibrated features are further employed to predict changed regions. The calibration in MOC is written as:
\begin{equation}
	\begin{aligned}
		\begin{array}{l}
			F_1^i = E_1^i - \hat E_2^i = E_1^i - E_2^i\left( {p{\rm{ + }}\Delta p} \right)\\
			F_2^i = E_2^i - \hat E_1^i = E_2^i - E_1^i\left( {p{\rm{ + }}\Delta p} \right)
		\end{array}
	\end{aligned}
\end{equation}
where ${p}$ enumerates the spatial coordinates in $E_1^i$ and $E_2^i$. $\Delta {\rm{p}}$ is the offset field of the corresponding coordinate. In order to realize adaptive alignment, the concatenated feature $E_C^i$ is fed into the offset estimation block (consist of two layers convolution and a channel shuffle operation) to compute the offset field $\Delta P$ between bi-temporal features. The degree of offset can be obtained by combining the vertical and horizontal dimensions, so the number of channels of $\Delta{P}$ is set to 2 and the spatial size of $\Delta {P}$ is the same as that of $E_1^i$ or $E_2^i$. Considering the integer attributes of spatial coordinates, the bilinear interpolation is utilized to adjust the offset values in $\Delta {P}$ as integers for calibration. Then, the features $E_1^i$ and $E_2^i$ are calibrated by:
\begin{equation}
	\begin{aligned}
		\begin{array}{l}
			\hat E_1^i\left( {\hat p} \right) = \sum\limits_q {{\cal B}\left( {\hat p,q} \right)E_1^i\left( q \right)} \\
			\hat E_2^i\left( {\hat p} \right) = \sum\limits_q {{\cal B}\left( {\hat p,q} \right)E_2^i\left( q \right)} 
		\end{array}
	\end{aligned}
\end{equation}
where ${\hat p = p + \Delta p}$ stands for the spatial position of sub-pixels in features $E_1^i$ and $E_2^i$. $q$ is the integer sampling grid at the corresponding spatial location. ${\cal B}$ denotes the bilinear interpolation operator and is defined as:
\begin{equation}
	\begin{aligned}
		\begin{array}{l}
			{\cal B}\left( {\hat p,{\kern 1pt} {\kern 1pt} {\kern 1pt} q} \right) = {\cal G}\left( {{{\hat p}_x},{q_x}} \right){\cal G}\left( {{{\hat p}_y},{q_y}} \right)\\
			{\cal G}\left( {c,d} \right) = Max\left( {0,1 - \left| {c - d} \right|} \right)
		\end{array}
	\end{aligned}
\end{equation}
where subscripts \emph{x} and \emph{y} are the horizontal and vertical directions, respectively. Through this module, the features from JCA modules are further enhanced. The feature mismatches in terms of spatial positions are suppressed efficiently owing to the introduction of offset estimation block.

\subsection{Feature Integration Module}
In order to efficiently compensate the spatial mismatches using calibrated features, we design the FeI module as shown in Fig. 2. In this module, the features at different scales are combined with the outputs of MOC modules. Specifically, the bi-temporal features at the \emph{i}th scale are both multiplied with the projection of ${D^i}$ to reduce the mismatches in these features. Then, they are concatenated together for the following CD prediction. The process of FeI module can be written as:
\begin{equation}
	\begin{aligned}
		{D^i} = \varphi \left( {C\left( {{I_1} + {I_1} \odot \varphi\left( {{F^i}} \right),{I_2} + {I_2} \odot \varphi\left( {{F^i}} \right),{F^i}} \right)} \right)
	\end{aligned}
\end{equation}
where ${{I_1}}$ and ${{I_2}}$ are the input bi-temporal features. $\varphi$ stands for the convolution block, and $C$  denote the concatenation operation.

\subsection{Optimization}
For the optimization of the suggested MSI-Net, we adopt the cross-entropy loss function. Considering that there is a sample imbalance between changing and unchanging pixels, the weighting mechanism is introduced within the cross-entropy loss, which is expressed as:
\begin{equation}
	\begin{aligned}
		\begin{array}{l}
			{L_{WCE}} =  - \sum {[{w_1}Y}  \odot \log \hat Y + {w_2}(1 - Y)  \odot \log (1 - \hat Y)]
		\end{array}
	\end{aligned}
\end{equation}
where $Y$ denotes the ground truth and $\hat Y$ is the predicted change result. ${{w_1}}$ and ${{w_2}}$ are set as 0.7 and 0.3, respectively. If the predicted change map value is 1, the pixel corresponding to that point is a change pixel, and vice versa.

\section{Experiments}
This section first describes two public datasets we used. Next, the constructed TUE-CD dataset for the assessment of building damage after earthquakes is described in detail. Then, we present all evaluation metrics, followed by a brief description of all comprised methods and the implementation details of the proposed MSI-Net. Finally, we analyze the results of on three considered datasets and verify the validity of MSI-Net.

\subsection{Description of Datasets}
In the experimental section, we conduct in-depth experiments of all methods on three datasets that contain distinct types of changes. These three datasets are WHU-CD \cite{Blpher45}, CLCD \cite{Blpher35}, and the proposed TUE-CD datasets. In these datasets, the WHU-CD dataset contains the changes of buildings, such as the emergence of buildings. The CLCD dataset focus on the changes of cropland. The proposed TUE-CD dataset records the building damage caused by earthquakes.

\emph{WHU-CD}: This dataset is composed of a pair of high-resolution aerial images measuring 32507 x 15345, containing three bands of R, G, B. In the following experiments, we crop the bi-temporal RS images into non-overlapping image patch with the size of 256×256. The numbers of images pairs in training, validation, and test datasets are 6069, 762, and 762, respectively.

\emph{CLCD}: This dataset includes 600 pairs of RS images collected by Gaofen-2, with the size of ${512 \times 512}$.These images, with a resolution of 1 m/pixel, record changes in land use patterns in parts of Guangzhou, China, from 2017 to 2019. In our implementation, these images are divided into 360/120/120 for training/validation/testing, respectively.

\emph{TUE-CD}: To efficiently estimate the damage after earthquakes, we construct a TUE-CD dataset, in which the bi-temporal RS images are obtained by the WorldView-2 satellite from Turkey. Specifically, a devastating magnitude 7.8 earthquake struck the southern part of Turkey bordering Syria at 4.17 a.m. local time on Monday, 6 February 2023. The epicenter of the earthquake is located approximately 23 kilometers east of Nurdaji, Gaziantep province (37.174°N, 37.032°E), which is near the Syrian border. These regions has experienced the strongest earthquakes in this century. The damage caused by this quake has exceeded 104 billion dollars. According to the public RS images from Maxar, 1656 image patches with the size of 256×256 are prepared and cover the most severely affected areas shown in Fig. 6, such as Adiyaman, Kahramanmaras, Hatay, and Gaziantep. In the TUE-CD dataset, 2338 destroyed buildings are labeled. In the following experiments, we divide the training, validation, and test sets according to the ratio of 7:1:2.

\subsection{Evaluation metrics}
In order to evaluate the CD performance of all the methods on the three datasets, we adopted the following five quantitative metrics as assessment: precision (P), recall (R), micro F1 score (mF1), mean intersection over unions (mIoU), and overall accuracy (OA). They are defined as follows:

\begin{equation}
	\begin{aligned}
		\begin{array}{c}
			\begin{gathered}
				{\rm{P}} = \frac{1}{N}\frac{{T{P_n}}}{{T{P_n} + F{P_n}}} \hfill \\
				\\
				{\rm{R}} = \frac{1}{N}\frac{{T{P_n}}}{{T{P_n} + F{N_n}}} \hfill \\
				\\
				{\rm{mF1}} = \frac{1}{N}\sum\limits_{n = 1}^N {\frac{{2 \times {P_n} \times {R_n}}}{{{P_n} + {R_n}}}}  \hfill \\
				\\
				{\rm{mIoU}} = \frac{1}{N}\sum\limits_{n = 1}^N {\frac{{T{P_n}}}{{T{P_n} + F{P_n} + F{N_n}}}}  \hfill \\
				\\
				{\rm{OA}} = \frac{{TP + TN}}{{TP + FP + TN + FN}} \hfill \\ 
			\end{gathered}
		\end{array}
	\end{aligned}
\end{equation}
where true positives (TP) and true negatives (TN) indicate the number of correctly predicted changed and unchanged pixels, respectively. The numbers of misestimated unchanged pixels and changed pixels are defined as false positives (FP) and false negatives (FN), respectively. P indicates the accuracy of detected regions containing change information and the regions without changes. Similarly, R refers to the recall of changed and unchanged regions, respectively. The mF1, mIoU and OA measure the overall performance of CD methods.

\subsection{Compared Methods and Implementation Details}
To demonstrate the effectiveness of the proposed MSI-Net, we conducted experiments comparing it with the following 9 state-of-the-art methods on three more datasets, comprising: FC-EF \cite{Blpher06}, FC-Siam-conc \cite{Blpher06}, FC-Siam-diff \cite{Blpher06}, dual task-constrained deep Siamese convolutional network (DTCDSCN) \cite{Blpher46}, bi-temporal image transformer (BIT) \cite{Blpher15}, global-aware siamese network (GAS-Net) \cite{Blpher13}, attention-based multi-scale transformer network  (AMTNet) \cite{Blpher47}, ultra lightweight spatial–spectral feature cooperation network (USSFC-Net) \cite{Blpher48}, and hierarchical attention network (HANet) \cite{Blpher49}. The characteristics of these methods are briefly described below:

\emph{FC-EF}: A DL-based CD method that concatenates multi-temporal RS images as the input of the network and employs a fully convolutional network.

\emph{FC-Siam-diff}: Multi-temporal RS image features are extracted separately by a Siamese fully convolutional network with shared weights, and then the absolute value of the inter-feature difference is used to infer the CD result.

\emph{FC-Siam-conc}: This method leverages a Siamese network structure and applies concatenated bi-temporal features for the inference of change regions.

\emph{DTCDSCN}: A Siamese network is enhanced by introducing a semantic segmentation task, in which change regions are highlighted with the assistance of spatial and channel attention mechanism.

\emph{BIT}: The network efficiently integrates CNN and transformer. In this method, semantic tokens are generated by segmentation, and cross-attention is employed to enhance CD performance.

\emph{GAS-Net}: The network analyzes the relationship between observed scene and foreground to produce effective global perceptual features and improves the robustness of the method.

\emph{AMTNet}: The network utilizes the attention mechanism to efficiently analyze the internal semantic information among inter-temporal RS images, and reduces the impact of domain gaps on CD by exchanging features from different images.

\emph{USSFC-Net}: The network applies multi-scale decoupled convolution to extract the representation information of the change region, and utilizes an potent spatial-spectral synergy strategy to acquire more discriminative features.

\emph{HANet}: The network combines a lightweight self-attention mechanism and a dense attention block to integrate multi-scale features, by which feature information is further refined.

\begin{figure*}[ht!]
	\centering	
	\includegraphics[scale=0.8]{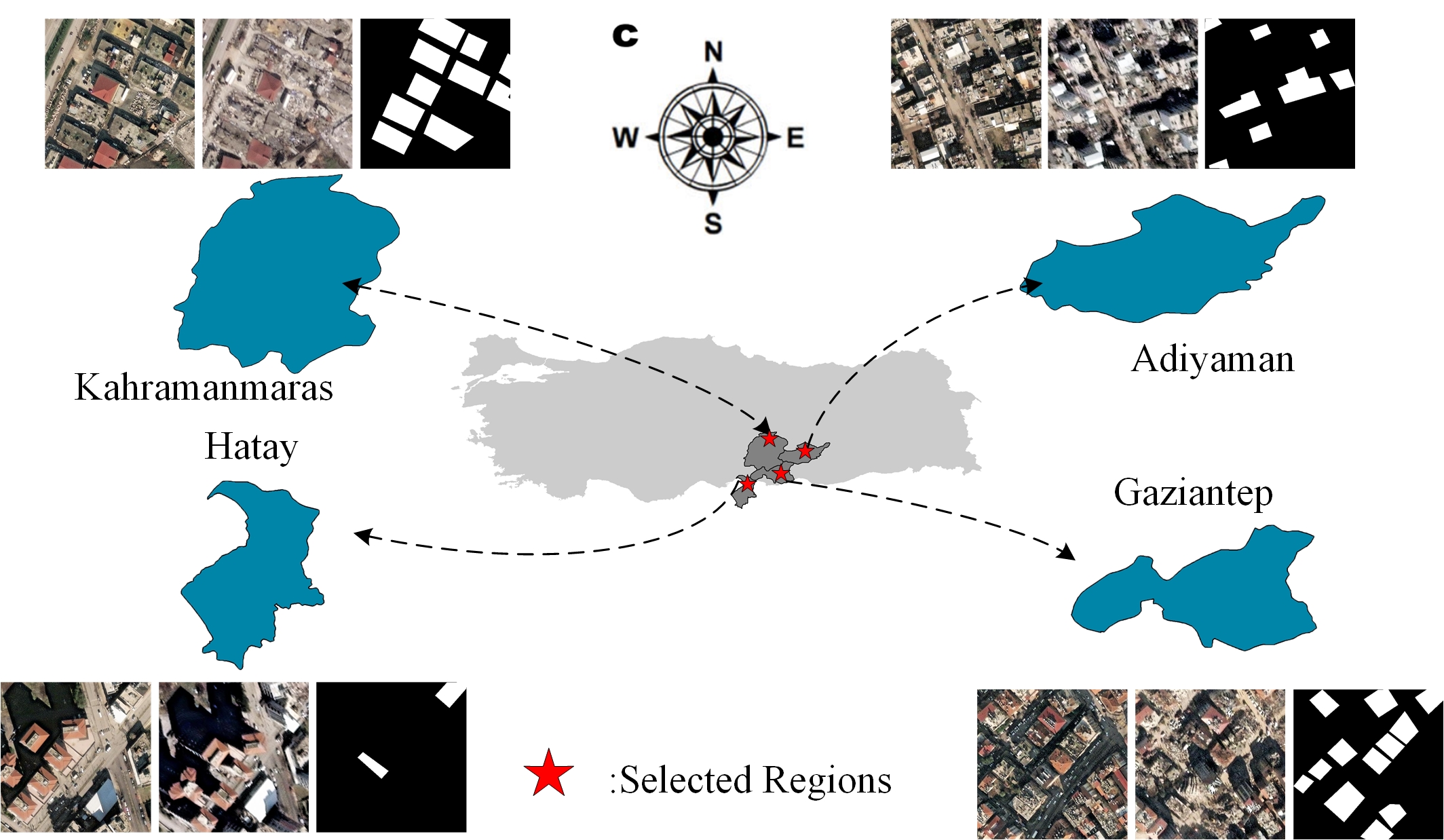}
	\caption{The most severely affected areas and some typical samples from these areas.}
	\label{fig.6}
\end{figure*}

\begin{figure*}[ht!]
	\centering	
	\includegraphics[scale=0.195]{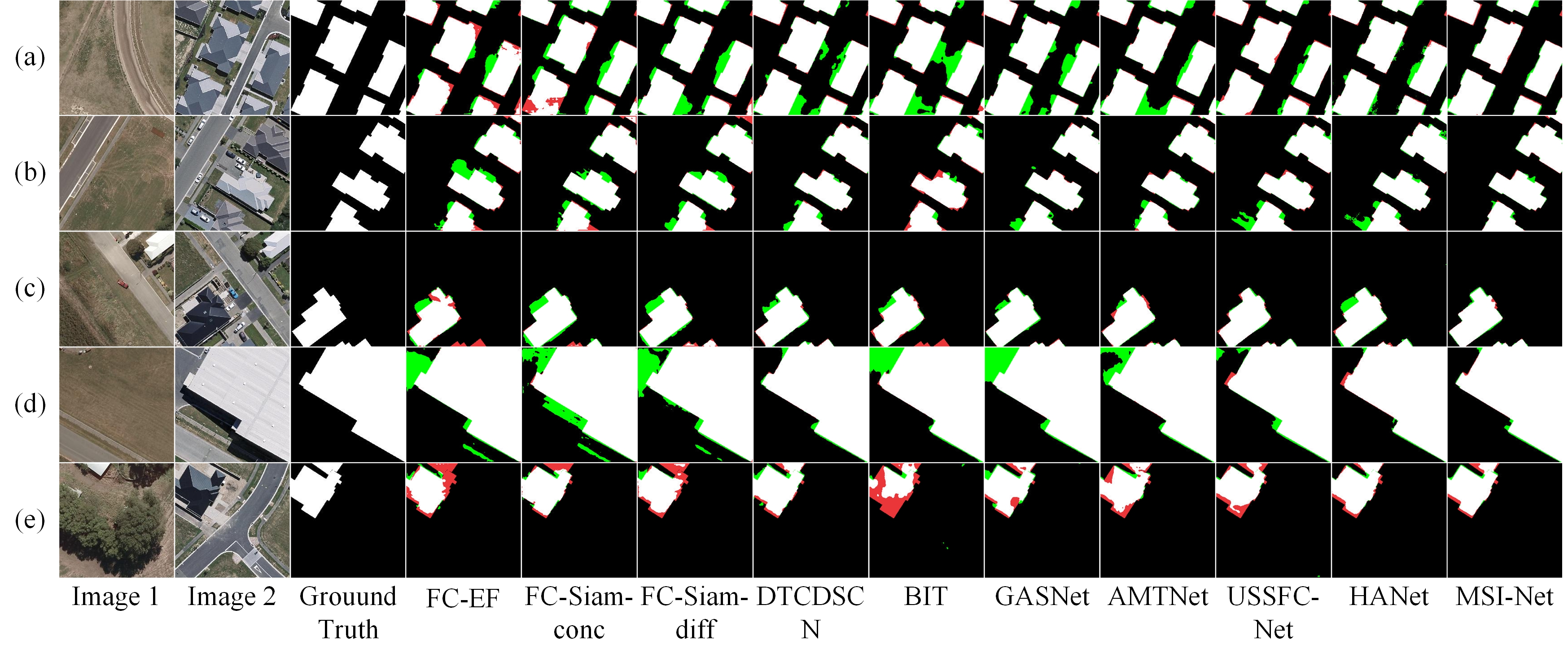}
	\caption{Qualitative comparison of all methods on the WHU-CD dataset. (a)–(e) Prediction results of all methods on different image pairs.}
	\label{fig.7}
\end{figure*}
The model parameters in our proposed MSI-Net are optimized by the AdamW optimizer. Based on the classical settings in other methods, we set the number of epochs to be trained to 200 and the initial learning rate to 0.001. In order to train MSI-Net more efficiently we adopted a linear recession strategy to gradually reduce the learning rate to 0 until all epochs are accomplished. In this paper, we set the batch size to 16, the proposed MSI-Net and all comparison methods adopt the same setting of training epochs and batch size.

\subsection{Comparison Performance with State-of-the-art Methods}
\emph{1) Comparison Experiments on the WHU-CD Dataset}: In this section the results of the CD experiments obtained on the WHU-CD dataset are discussed. Table 1 shows the evaluation metrics achieved by MSI-Net and the considered comparative methods on this dataset. The best values are the ones marked in bold. As can be seen from the Table 1, the proposed MSI-Net outperforms other methods in terms of P, mF1, mIoU and OA with 96.66\%, 95.58\%, 91.81\% and 99.27\%, respectively. In addition, the R value of our proposed MSI-Net achieves the second best result, which is only 0.02\% lower than the GAS-Net. Overall, these metrics reflect the effectiveness of our method. Therefore, MSI-Net achieved the best results on more metrics and better performance than the comparison methods.

To demonstrate the validity of the proposed MSI-Net on the WHU-CD dataset more intuitively, we select several samples and visualized their CD map in Fig. 7. In order to more intuitively observe the accuracy of the CD results, we denote TP, TN, FP, and FN in white, black, red, and green, respectively. As shown in Figs. 7(a), 7(b), 7(d), many false positive regions (labeled in green) appear in the CD results of FC-EF, FC-Siam-diff, FC-Siam-conc, DTCDSCN, BIT, GAS-Net, and AMTNet. In Fig. 7(c), all methods except USSFC-Net and the proposed MSI-Net wrongly recognize the unchanged parts of roads as changed areas. In Fig. 7(e), all compared methods suffer from the blurring of edge information and fail to detect the changed region completely due to the image mismatch problem. In contrast, the proposed MSI-Net can reduce the influences of the side-looking issue on the CD accuracy by introducing MOC modules and obtain more complete edge information.

\begin{figure*}[ht!]
	\centering	
	\includegraphics[scale=0.195]{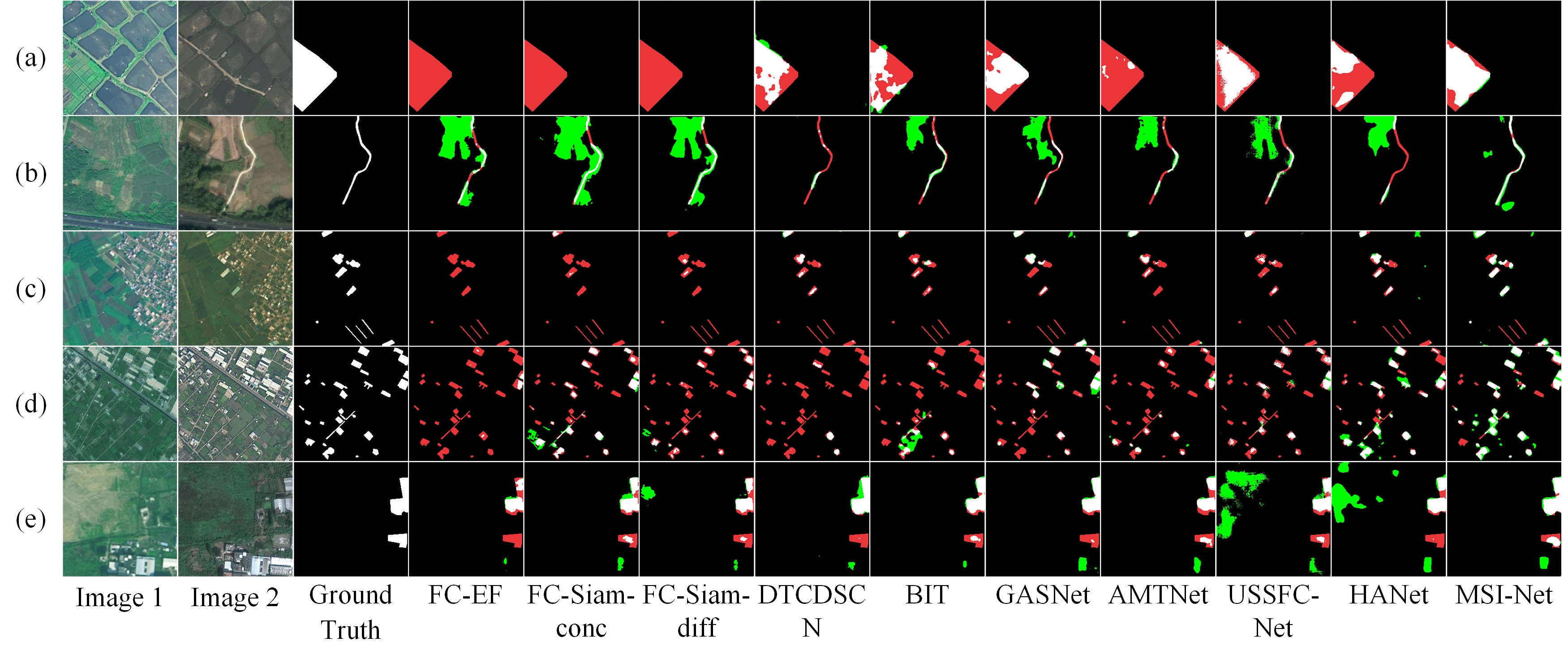}
	\caption{Qualitative comparison of all methods on the CLCD dataset. (a)–(e) Prediction results of all methods on different image pairs.}
	\label{fig.8}
\end{figure*}

\begin{figure*}[ht!]
	\centering	
	\includegraphics[scale=0.195]{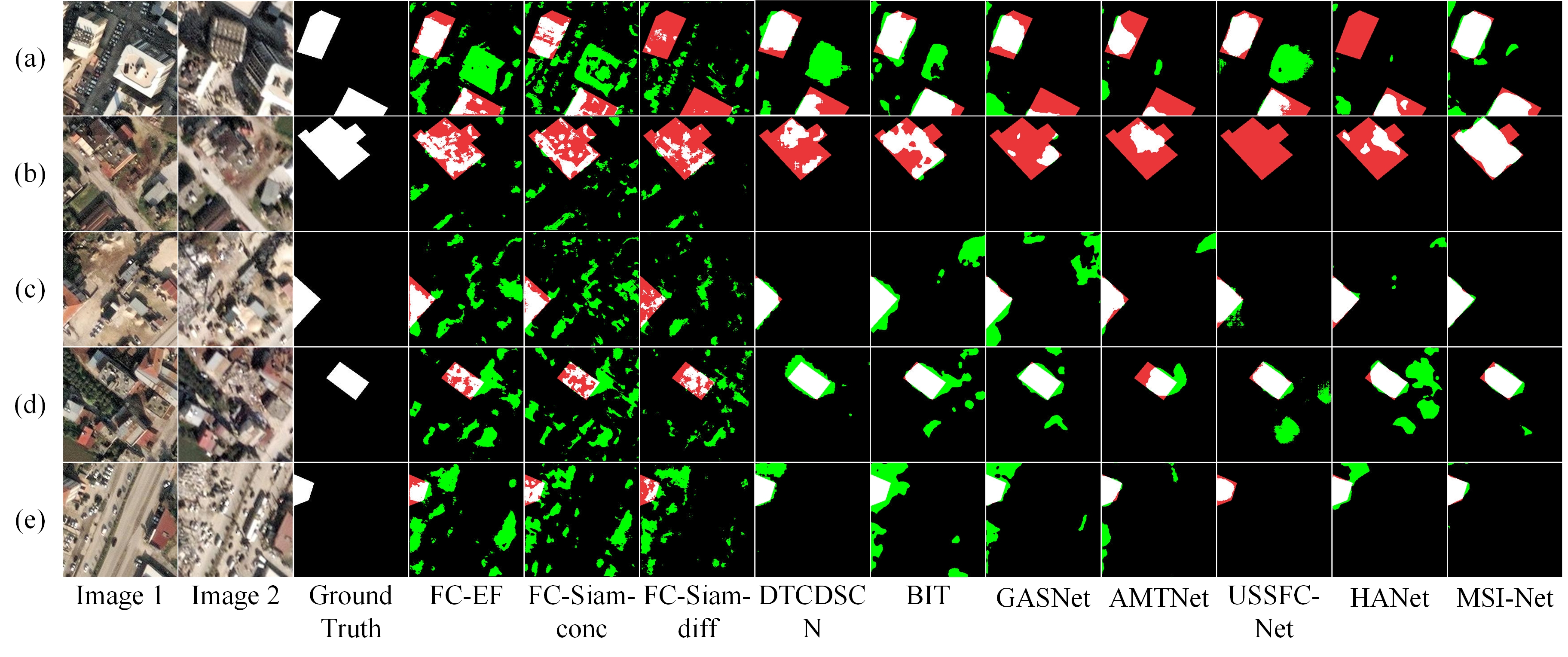}
	\caption{Qualitative comparison of all methods on the TUE-CD dataset. (a)–(e) Prediction results of all methods on different image pairs.}
	\label{fig.9}
\end{figure*}

\begin{table}[h]
	\centering
	\captionsetup{font={footnotesize},skip=1pt}
	\setlength\tabcolsep{4.5pt}
	%    \captionsetup{font={scriptsize}}
	\caption{Quantitative results for all methods on the WHU dataset.}
	\footnotesize
	%   \scriptsize
	\renewcommand\arraystretch{1.3}
	\begin{tabular}{cccccc}
		\hline
		Metric (\%)  & P              & R              & mF1             & mIoU            & OA             \\ \hline
		FC-EF        & 93.20          & 90.21          & 91.64          & 85.50          & 98.64          \\
		FC-Siam-conc & 89.17          & 91.63          & 90.36          & 83.59          & 98.33          \\
		FC-Siam-diff & 79.71          & 92.22          & 87.60          & 79.71          & 97.66          \\
		DTCDSCN      & 94.96          & 88.75          & 91.60          & 85.41          & 98.48         \\
		BIT       & 93.31 & 94.46          & 93.90          & 89.01          & 98.96          \\
		GAS-Net       & 90.16          & \textbf{94.57}          & 92.24          & 86.39          &  98.62  \\
		AMTNet       & 93.36          & 92.20          & 92.77          & 88.81          &  98.94  \\
		USSFC-Net          & 96.60          & 92.77          & 94.59          &90.16           & 99.13          \\
		HANet & 95.21          & 94.55          & 94.88          & 90.63          & 99.14          \\
		MSI-Net     & \textbf{96.66}          & 94.55 & \textbf{95.58} & \textbf{91.81} &  \textbf{99.27} \\ \hline
	\end{tabular}
	\label{tableI}
	\vspace{-0.2cm}
\end{table}
\begin{table}[h]
	\centering
	\captionsetup{font={footnotesize},skip=1pt}
	\setlength\tabcolsep{4.5pt}
	%    \captionsetup{font={scriptsize}}
	\caption{Quantitative results for all methods on the CLCD dataset.}
	\footnotesize
	%   \scriptsize
	\renewcommand\arraystretch{1.3}
	\begin{tabular}{cccccc}
		\hline
		Metric (\%)  & P              & R              & mF1             & mIoU            & OA             \\ \hline
		FC-EF        & 77.93          & 65.03          & 70.90          & 61.54          & 94.38          \\
		FC-Siam-conc & 85.50          & 73.15          & 77.77          & 67.92          & 94.93          \\
		FC-Siam-diff & 81.06          & 78.36          & 79.64          & 69.74          & 94.63          \\
		DTCDSCN      & 80.56          & 77.75          & 79.07          & 69.12          & 94.50         \\
		BIT       & 80.63 & 75.19          & 77.59          & 67.59          & 94.37          \\
		GAS-Net       & \textbf{85.65}          & 77.02          & 80.51          & 70.57          &  94.25  \\
		AMTNet       & 83.73          & 77.72          & 80.37          & 70.62          &  95.07  \\
		
		USSFC-Net          & 85.21          & 79.75          & 82.20          &72.72           & 95.26          \\
		HANet & 83.30          & 81.47          & 82.36          & 72.85          & 95.20          \\
		MSI-Net     & 84.21          & \textbf{83.30} & \textbf{82.96} & \textbf{73.53} &  \textbf{95.47} \\ \hline
	\end{tabular}
	\label{tableII}
	\vspace{-0.2cm}
\end{table}	

\begin{table}[h]
	\centering
	\captionsetup{font={footnotesize},skip=1pt}
	\setlength\tabcolsep{4.5pt}
	%    \captionsetup{font={scriptsize}}
	\caption{Quantitative results for all methods on the TUE-CD dataset.}
	\footnotesize
	%   \scriptsize
	\renewcommand\arraystretch{1.3}
	\begin{tabular}{cccccc}
		\hline
		Metric (\%)  & P              & R              & mF1             & mIoU            & OA             \\ \hline
		FC-EF        & 59.13          & 67.15          & 61.60          & 52.94          & 89.10          \\
		FC-Siam-conc & 58.17          & 67.14          & 60.32          & 51.96          & 88.71          \\
		FC-Siam-diff & 56.53          & 61.62          & 57.98          & 50.67          & 89.34          \\
		DTCDSCN       & 75.48          & 71.26         & 73.15          & 63.78          &  93.53  \\
		BIT       & 75.28 & 75.91          & 76.50          & 66.83          & 95.52          \\
		GAS-Net       & 74.25          & 75.91          & 75.05          & 65.42          &  95.32  \\
		AMTNet      & 75.90          & \textbf{77.99}          & 76.89          & 67.24          & 95.64         \\
		USSFC-Net          & \textbf{80.99}          & 72.44          & 75.94          &66.50           & 95.38          \\
		HANet & 78.60          & 74.96          & 76.63          & 67.10          & 96.03          \\
		MSI-Net     & 79.17          & 76.97 & \textbf{78.02} & \textbf{68.48} &  \textbf{96.16} \\ \hline
	\end{tabular}
	\label{tableIII}
	\vspace{-0.2cm}
\end{table}

\begin{figure*}[ht!]
	\centering	
	\includegraphics[scale=0.3]{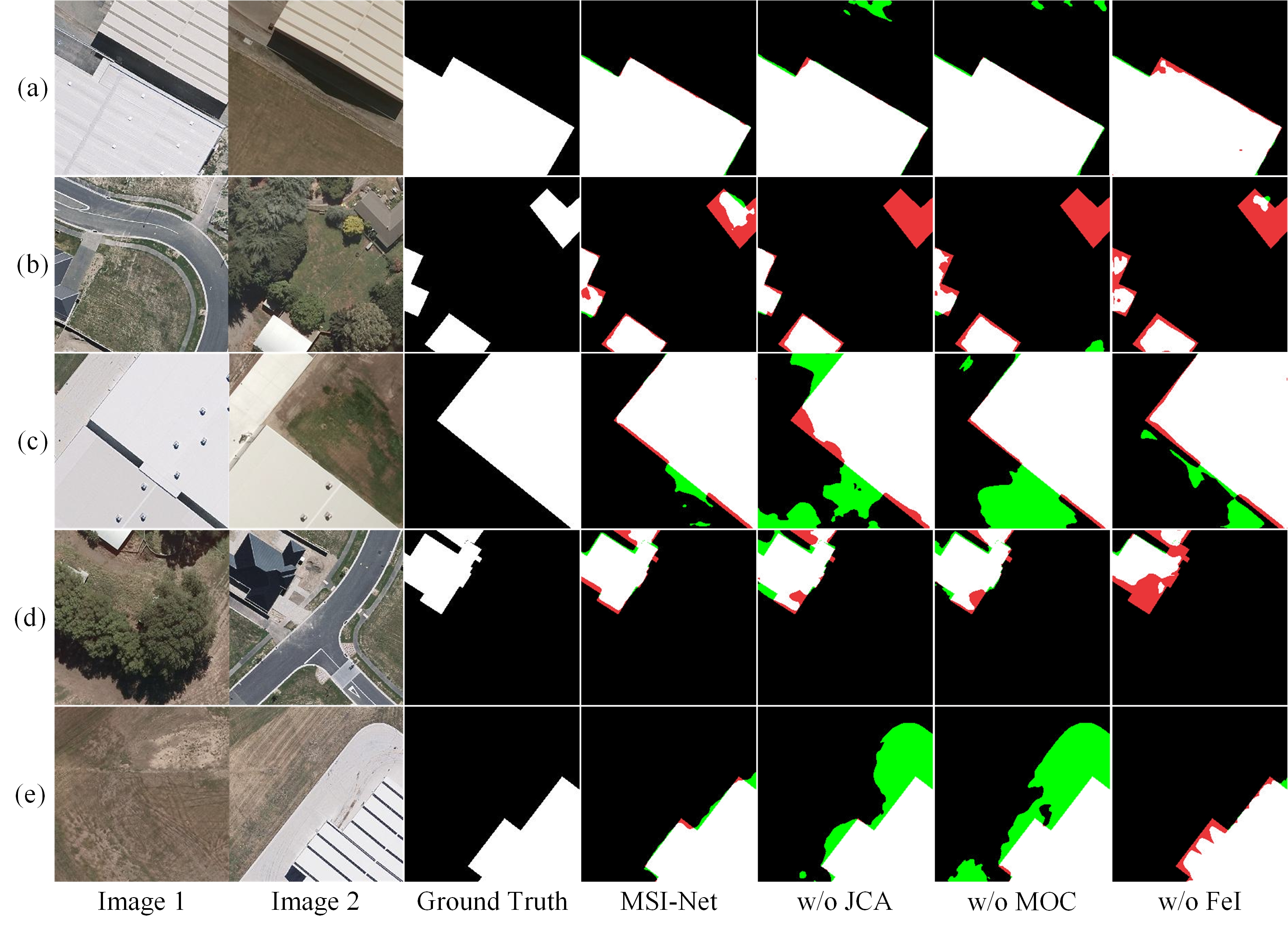}
	\caption{Examples of qualitative results of the ablation study on the WHU-CD dataset. (a)–(e) Prediction results of all methods on different image pairs.}
	\label{fig.10}
\end{figure*}
\begin{figure*}[ht!]
	\centering	
	\includegraphics[scale=0.3]{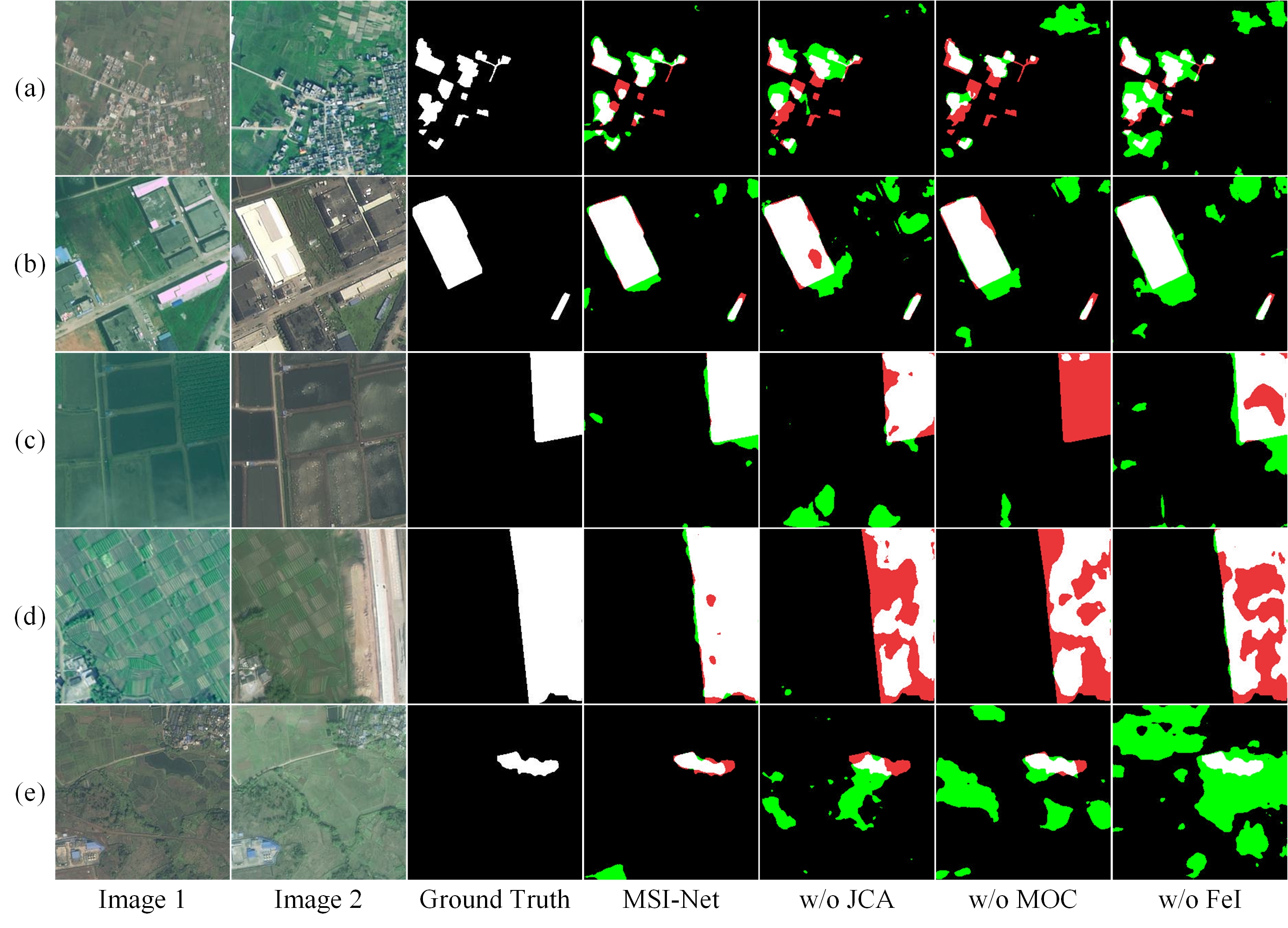}
	\caption{Examples of qualitative results of the ablation study on the CLCD dataset. (a)–(e) Prediction results of all methods on different image pairs.}
	\label{fig.11}
\end{figure*}
\begin{figure*}[ht!]
	\centering	
	\includegraphics[scale=0.3]{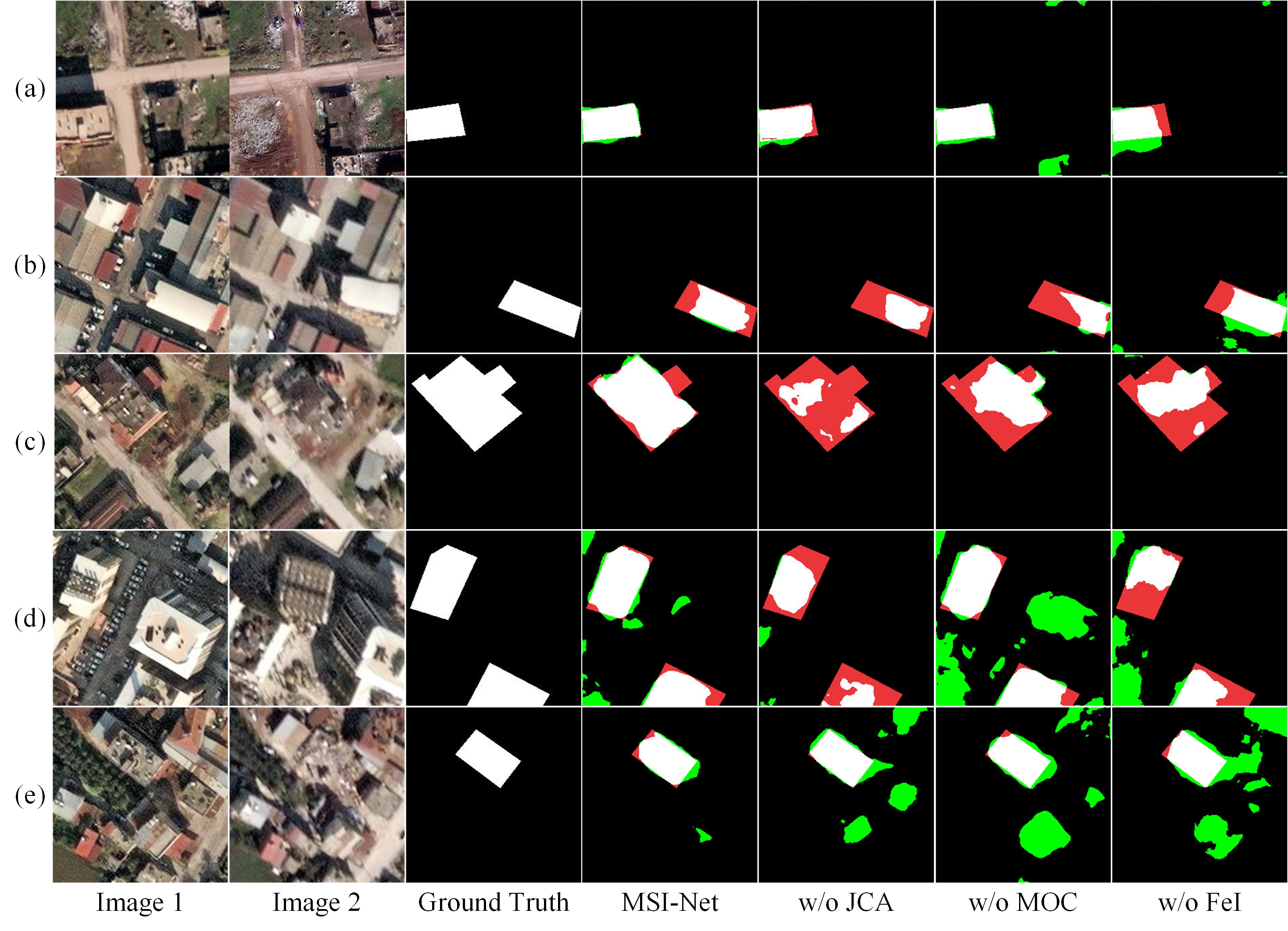}
	\caption{Examples of qualitative results of the ablation study on the TUE-CD dataset. (a)–(e) Prediction results of all methods on different image pairs.}
	\label{fig.12}
\end{figure*}

\emph{2) Comparison Experiments on the CLCD Dataset}: The results obtained by MSI-Net and all comparison methods on the CLCD dataset are shown in Table 2. This dataset records cropland utilization changes with a lower resolution and involves more complex scenarios. From this table, it can be noticed that the suggested MSI-Net achieves the best R, mF1, mIoU, OA. Moreover, the mF1 is 0.6\% higher than HANet which produces the second-best value. This implies that the proposed MSI-Net is more efficient in terms of complex scenarios compared to other methods because of the introduction of the joint attention.

Fig. 8 shows the visualization results of some typical samples in the CLCD dataset. In Fig. 8(a), it can be seen that the proposed MSI-Net can obtain a more completed change region compared to other methods. In Fig. 8(b), the result of the proposed MSI-Net contains more completed roads, in addition to having fewer false positive regions. As shown in Figs. 8(c), (d), the proposed MSI-Net is able to find change regions with small sizes in complex scenarios. This is primarily attributed to the fact that the designed JCA module can model the long-range dependencies between different images in both channel and spatial dimensions.

\emph{3) Comparison Experiments on the TUE-CD Dataset}: In Table 3, the metrics obtained by all methods for each evaluation indicator are presented. The USSFC-Net and AMTNet produces the best P and R, respectively. The second-best P and R are both from our proposed MSI-Net. Besides, MSI-Net provides the best mF1, mIoU, and OA. Since the image pairs in the TUE-CD dataset contain complex scenarios and suffer from severe side-looking problems, it is more difficult to perform CD. But, the values of mF1 and mIoU achieved by the proposed MSI-Net are 1.13\% and 1.24\% higher than those of AMTNet, respectively. Therefore, it can be deduced that the proposed MSI-Net is more effective in suppressing the effect of the side-looking problem on the CD results on this dataset.

Fig. 9 displays some CD results of all methods on the TUE-CD dataset. In the TUE-CD dataset, collapsed buildings are labeled. From Fig. 9(a), it can be found that there is a side-looking problem in terms of building areas of multi-temporal RS images. From Figs. 9(a) and 9(d), it can be observed that the results of all comparison methods show many false positive regions. In addition, there are some large misestimated areas in the results of FC-EF, FC-Siam-conc, and FC-Siam-diff. In Figs. 9(c) (e), the proposed method can locate the complete regions of collapsed buildings with better edge information and less false positive regions. This is mainly due to the fact that the designed MOC module can alleviate the influence of the side-looking problem on the CD accuracy.

\begin{table}[]
	\centering
	\captionsetup{font={footnotesize},skip=1pt}
	\setlength\tabcolsep{4.5pt}
	%    \captionsetup{font={scriptsize}}
	\caption{Ablation study of JCA, MOC, and FeI modules on WHU-CD, CLCD, and TUE-CD datasets.}
	\footnotesize
	%   \scriptsize
	\renewcommand\arraystretch{1.3}
	\begin{tabular}{cllllll}
		\hline
		Datasets                 & Settings & P     & R     & F1    & IoU   & OA    \\ \hline
		\multirow{4}{*}{WHU-CD}   & w/o JCA  & 90.12 & \textbf{95.25} & 92.61 & 86.81 & 98.66 \\
		& w/o MOC  & 90.59 & 94.65 & 92.58 & 86.84 & 98.68 \\
		& w/o FeI & \textbf{97.18} & 92.77 & 94.86 & 90.60 & 99.18 \\
		& MSI-Net & 96.66 & 94.55 & \textbf{95.58} & \textbf{91.81} & \textbf{99.27} \\ \hline
		\multirow{4}{*}{CLCD} & w/o JCA  & 78.99 & 77.47 & 78.20 & 68.15 & 94.15 \\
		& w/o MOC  & 80.62 & 76.74 & 78.52 & 68.54 & 94.45 \\
		& w/o FeI & 83.06 & 81.30 & 82.15 & 72.61 & 95.21 \\
		& MSI-Net & \textbf{84.21} & \textbf{83.30} & \textbf{82.96} & \textbf{73.53} & \textbf{95.26} \\ \hline
		\multirow{4}{*}{TUE-CD}    & w/o JCA  & 79.12 & 74.39 & 76.51 & 66.99 & 96.08 \\
		& w/o MOC  & 74.31 & \textbf{78.92} & 76.38 & 66.66 & 95.31 \\
		& w/o FeI & 75.73 & 78.59 & 77.07 & 67.41 & 95.61 \\
		& MSI-Net & \textbf{79.17}& 76.97 & \textbf{78.02} & \textbf{68.48} & \textbf{96.16} \\ \hline
	\end{tabular}
	\label{tableIV}
	\vspace{-0.3cm}
\end{table}
\begin{table}[]
	\centering
	\captionsetup{font={footnotesize},skip=1pt}
	\setlength\tabcolsep{4.5pt}
	%    \captionsetup{font={scriptsize}}
	\caption{Analysis of the CCA, and SJA on WHU-CD, CLCD, and TUE-CD datasets.}
	\footnotesize
	%   \scriptsize
	\renewcommand\arraystretch{1.3}
	\begin{tabular}{cllllll}
		\hline
		Datasets                 & Settings & P     & R     & F1    & IoU   & OA    \\ \hline
		\multirow{4}{*}{WHU-CD}   & w/o CCA  & 87.96 & 95.03 & 91.16 & 84.75   & 98.37\\
		& w/o SJA & 88.96 & \textbf{96.11} & 92.20 & 86.32 & 98.57 \\
		& MSI-Net & \textbf{96.66} & 94.55 & \textbf{95.58} & \textbf{91.81} & \textbf{99.27} \\ \hline
		\multirow{4}{*}{CLCD} & w/o CCA  & 78.33 & 78.24 & 78.29 & 68.20 & 94.03 \\
		& w/o SJA & 76.46 & 81.79 & 78.51 & 68.27 & 93.41 \\
		& MSI-Net & \textbf{84.21} & \textbf{83.30} & \textbf{82.96} & \textbf{73.53} & \textbf{95.26} \\ \hline
		\multirow{4}{*}{TUE-CD}    & w/o CCA  & \textbf{79.62} & 73.17 & 75.95 & 66.47 & 95.73 \\
		& w/o SJA  & 76.23 & 75.49 & 75.86 & 66.26 & 95.68 \\
		& MSI-Net & 79.17 & \textbf{76.97} & \textbf{78.02} & \textbf{68.48} & \textbf{96.16} \\ \hline
	\end{tabular}
	\label{tableV}
	\vspace{-0.3cm}
\end{table}

\subsection{Ablation Study}
In order to analyze the validity of the various modules designed in MSI-Net, we perform ablation experiments of JCA, MOC, and FeI modules on the above three datasets. As shown in Table 4, we examine the validity of JCA, MOC, and FeI modules, respectively. In this table, w$/$o denotes that its corresponding module is ablated from MSI-Net. On the WHU-CD dataset, when the FeI module is ablated, the obtained results exhibit a small increase in terms of P by 0.52\%, while R is down by 1.78\%. Meanwhile, its ablation also results in the decrease of mF1 and mIoU. When the MOC module is eliminated, the metric R increases slightly but other metric values decrease dramatically. The reason for this is that the side-looking problems in bi-temporal images have negative influences on the CD results. In the CLCD dataset, the proposed MSI-Net achieves the best results in terms of all metrics. This proves that the designed JCA, MOC, and FeI modules can improve the performance of CD effectively on this dataset. For the TUE-CD dataset, the best P, mF1, mIoU, and OA are from the complete MSI-Net. The R increases with the removal of the MOC module, but the overall results of the complete MSI-Net perform better.

Figs. 10-12 show the results of ablation experiments of the proposed MSI-Net on three considered datasets. From Figs. 10(b) and 10(d), it can be noticed that when the MOC module is removed, the edges of changed regions are blurred and incomplete. So, the influences of mismatches on CD results are more obvious. Meanwhile, the CD results in Fig. 10(e) indicate that MSI-Net without JCA modules incorrectly estimate the irrelevant areas as changed regions. Fig. 11(e) demonstrates that removing the FeI module leads to a large green area in the CD results. This is probably because the features of multi-temporal images cannot be integrated together efficiently to reduce the impact of pseudo-change regions. So, many false positive regions are produced. From Fig. 12(c), it can be observed that when the JCA module is erased, the network is unable to capture the entire change region.
\begin{figure*}[ht!]
	\centering	
	\includegraphics[scale=0.35]{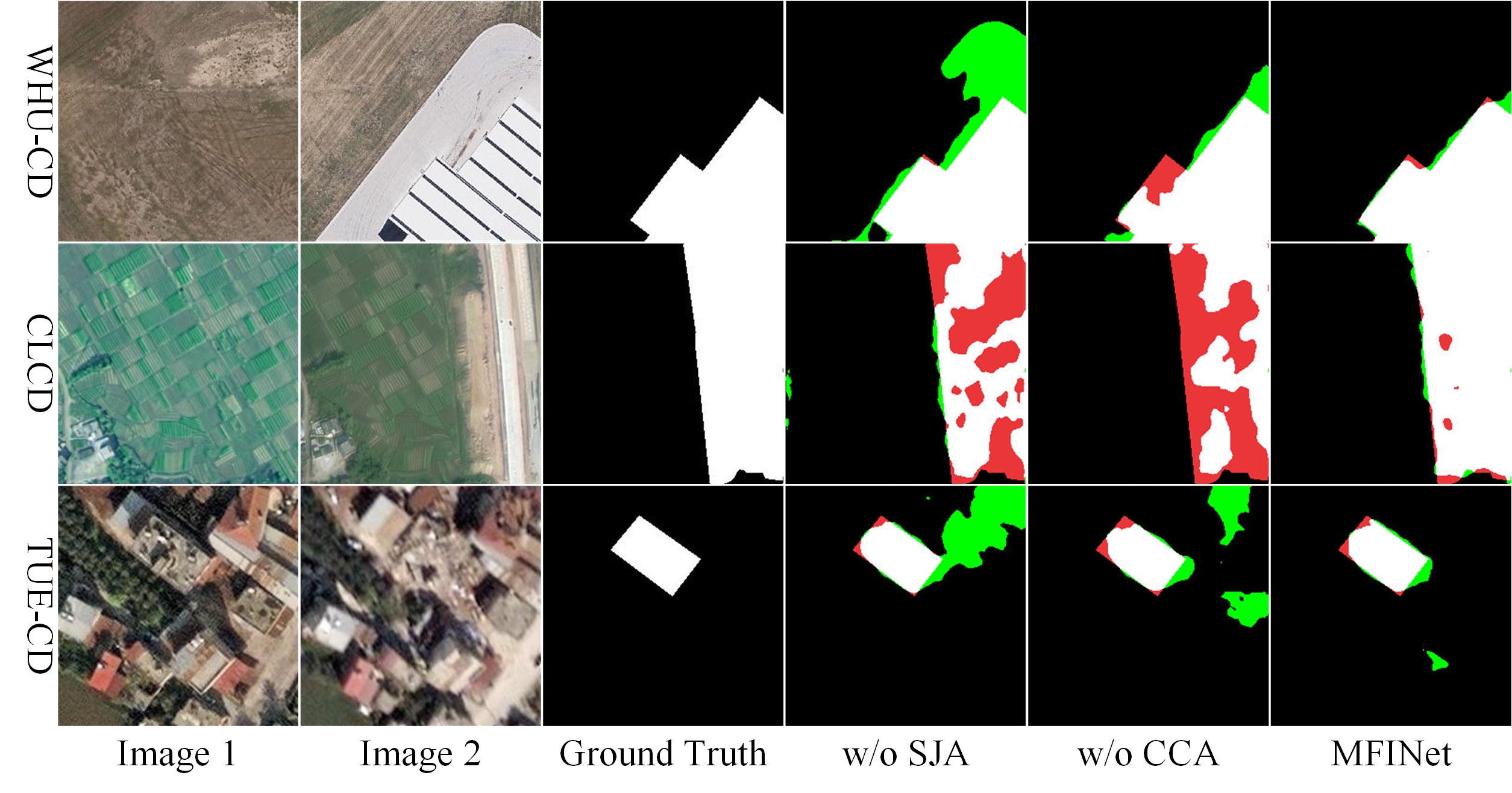}
	\caption{Qualitative results of different network configurations on the three datasets.}
	\label{fig.13}
\end{figure*}
\begin{figure*}[ht!]
	\centering	
	\includegraphics[scale=0.35]{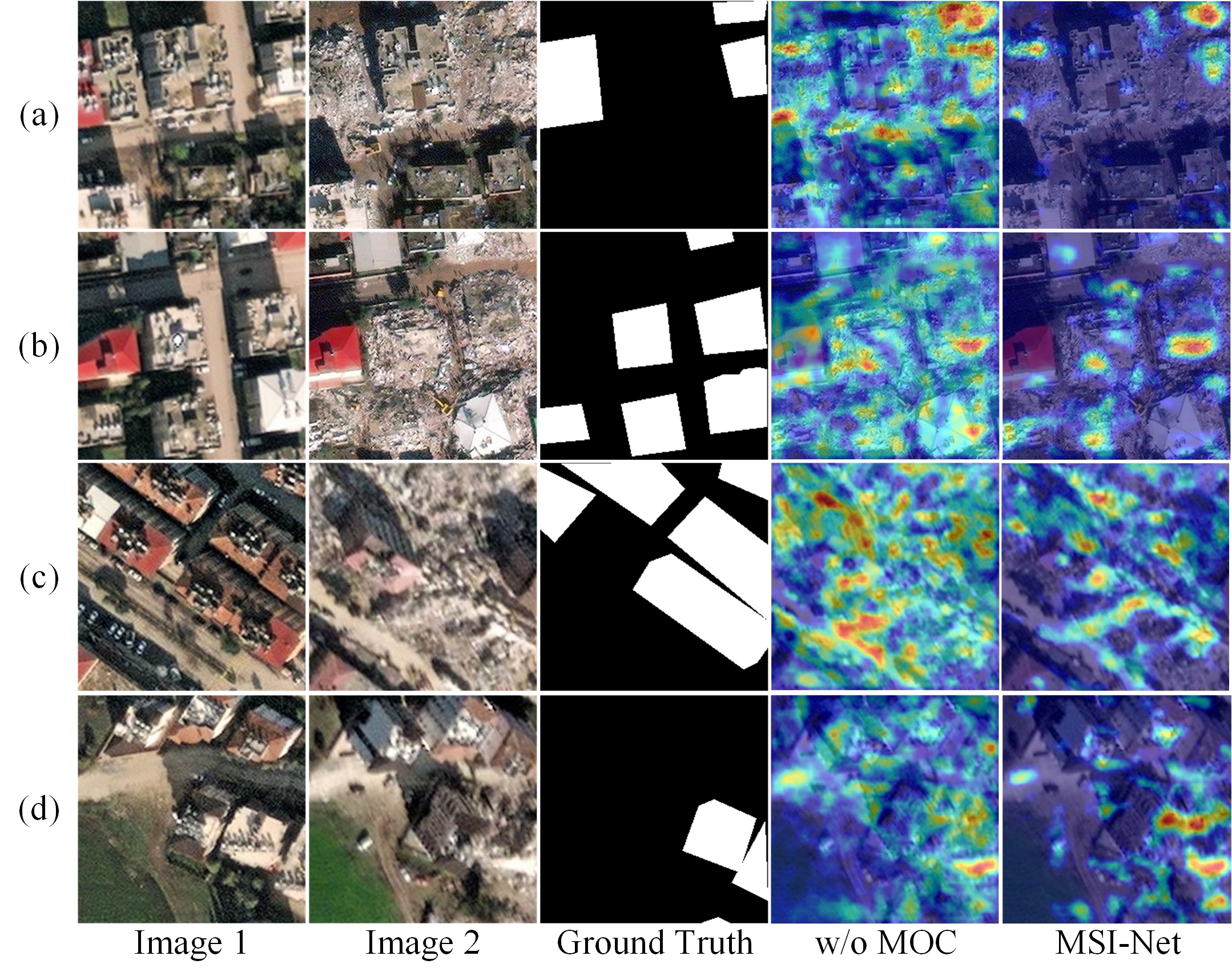}
	\caption{Feature visualization results of different method on the TUE-CD dataset.}
	\label{fig.14}
\end{figure*}

% Please add the following required packages to your document preamble:
% \usepackage{multirow}

\subsection{Analysis of of CCA and SJA}
There are two blocks CCA and SJA in JCA modules, which extract features of changed areas along channel and spatial dimensions. To verify the effectivenss of CCA and SJA blocks, we explored the effectiveness of these two blocks on the three datasets mentioned above. Table 5 exhibits the ablation results of the two blocks on three datasets. From this table, we can observe that there are some declines in terms of all evaluation values on the WHU-CD dataset when the CCA block is removed. For the ablation of the SJA block, the value of R increases, but other metrics suffer from a performance degradation. For the CLCD dataset, all metric values become worse when CCA or SJA block is erased. Besides, the evaluation values on the TUE-CD dataset demonstrate that eliminating CCA and SJA blocks negatively impacts CD performance. So, the introductions of CCA and SJA blocks can improve the overall CD accuracy.

Fig. 13 exhibits the CD results of the above ablation experiments. From the first and third rows in Fig. 13, it can be noticed that when the SJA block is removed, more green areas appear in the results. The reason for this is that spatial details in multi-temporal RS images cannot be extracted by MSI-Net without SJA blocks. From the second row of Fig. 13, we can noticed that the detected region become incomplete when CCA and SJA blocks are removed. Therefore, the complete MSI-Net obtains better results because of the configured CCA and SJA blocks.

\subsection{Feature Visualization }
For better visualization of the effectiveness of the MOC module in addressing the mis-matching problem in the TUE-CD dataset caused by the side-view problem, we visualized the output of the MOC module with the assistance of Grad-CAM. Fig. 14 shows the visualization  results of the ablation experiments on the MOC module, specifically "w$/$o MOC" means that we removed the MOC module and replaced it with a concat operation. Based on Fig. 14(c)(d), it is evident that taller buildings experience more significant side-looking issues, which have a greater impact on CD. It can be noticed that, the feature maps processed by the MOC module are more focused on the real changed areas than the feature maps without the MOC module.
\begin{figure*}[ht!]
	\centering	
	\includegraphics[scale=0.35]{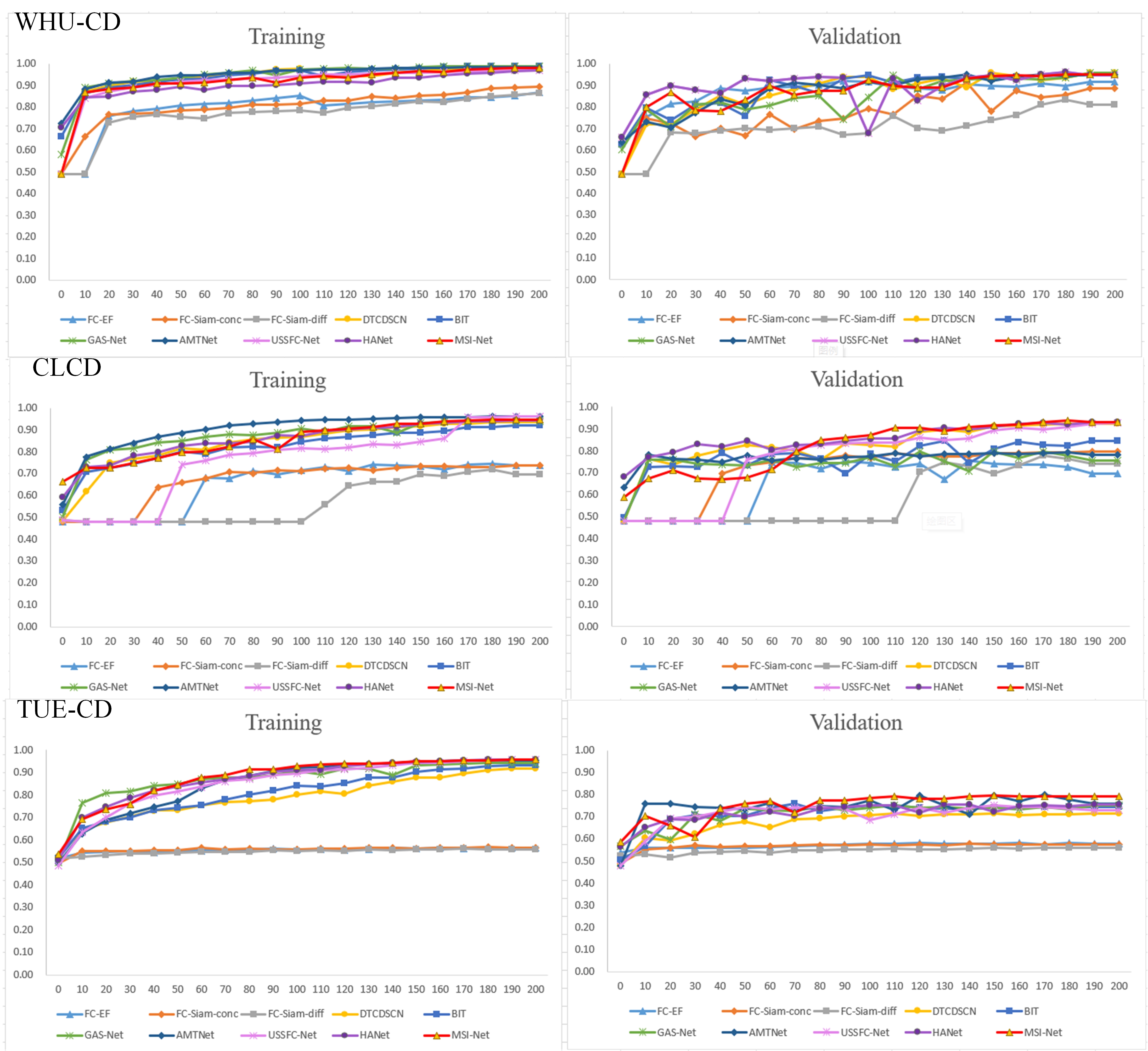}
	\caption{mF1-score of all methods for each epoch on the WHU-CD datasets}
	\label{fig.15}
\end{figure*}
\subsection{Training Convergence}
The training and validation process of all methods on the three datasets is shown in Fig. 15, where the vertical axis denotes the $\rm{m{F_1}}$ scores the horizontal axis denotes the number of epoach trained. The first and second row of fig. 15 demonstrate that all methods achieved a similar growth trend in mF1 during training on WHU-CD and CLCD datasets, with some fluctuations in the validation results. However, during training on the proposed TUE-CD dataset, both FC-EF, FC-Siam-conc, and FC-Siam-diff show a slight increase in mF1, as observed in the third row of Fig. 15. This could be due to the fact that these three methods based on full convolution are too simple for the task of CD in complex scenarios. The proposed MSI-Net grows more steadily and reached more competitive capabilities than the other comparison methods. As can be seen in Fig. 15, the mF1-score tends to converge when the number of training epochs reaches 200, while the verification performance does not improve more significantly. Consequently, we set the number of training epoach to 200.

\section{Conclusion}
\label{Conclusions}
In this work, we designed MSI network for RS image change detection. For MSI-Net, the JCA module extracts the long-range dependencies and boost the information exchange between bi-temporal features. Moreover, in order to mitigate the influence of mismatches caused by side-looking issues on CD accuracy, we propose the MOC module to calibrate the offsets. In addition, we construct a CD dataset, TUE-CD, for the assessment of building damage in the short term after an earthquake. The effectiveness of the suggested MSI-Net is verified on WHU-CD, CLCD, and the labeled TUE-CD dataset. Compared to the other nine state-of-the-art CD methods, MSI-Net perform better in terms of qualitative and quantitative evaluation results.

\bibliographystyle{IEEEtran}
\bibliography{Reference}

\end{document}